\documentclass{svproc}
\usepackage{graphicx}
\usepackage[autostyle]{csquotes}
\usepackage[english]{babel}
\usepackage[table]{xcolor}
\usepackage{tabu}
\usepackage[utf8]{inputenc}
\usepackage{array}
\usepackage{multirow}
\usepackage[ruled,linesnumbered]{algorithm2e}
\usepackage{url}

\begin{document}
\mainmatter   

\title{Toward Robust Image Classification}

\titlerunning{Toward Robust Image Classification} 

\author{Basemah Alshemali\inst{1}\inst{2} \and Alta Graham\inst{2} \and Jugal Kalita\inst{2}}

\authorrunning{B Alshemali, A Graham, and J Kalita}

\institute{ Computer Science Department, 
College of Computer Science and Engineering, 
Taibah University, 
Almadinah, KSA
\and 
Computer Science Department, 
College of Engineering and Applied Science, 
University of Colorado at Colorado Springs, 
Colorado Springs, USA \\
\email{
\{balshema,agraham,jkalita\}@uccs.edu
}
}

\maketitle

\begin{abstract}
Neural networks are frequently used for image classification, but can be vulnerable to misclassification caused by adversarial images. 
Attempts to make neural network image classification more robust have included variations on preprocessing (cropping, applying noise, blurring), adversarial training, and dropout randomization. 
In this paper, we implemented a model for adversarial detection based on a combination of two of these techniques: dropout randomization with preprocessing applied to images within a given Bayesian uncertainty. 
We evaluated our model on the MNIST dataset, using adversarial images generated using Fast Gradient Sign Method (FGSM), Jacobian-based Saliency Map Attack (JSMA) and Basic Iterative Method (BIM) attacks. 
Our model achieved an average adversarial image detection accuracy of 97\%, with an average image classification accuracy, after discarding images flagged as adversarial, of 99\%. 
Our average detection accuracy exceeded that of recent papers using similar techniques.
\keywords{Adversarial examples, Robust image classification, Bayesian uncertainty, Image cropping.}
\end{abstract}


\section{Introduction}
Deep neural networks (DNN) produce excellent image classification results and are the current state-of-the-art, but have been shown to be vulnerable to attacks by adversarial examples: Images altered by the introduction of small perturbations that cause the neural network to misclassify the image \cite{carlini2017adversarial}. 
\par
Many researchers have proposed defenses against adversarial image attacks on neural network classification systems. 
In image preprocessing defenses, the images are altered in some way before being classified (blurring, cropping, noise) in order to disrupt any adversarial perturbations \cite{graese2016assessing} \cite{guo2017countering}. 
In dropout randomization defenses, as the name suggests, the neural network adds randomization which supports the use of Bayesian uncertainty measurements to assess the likelihood of an image being adversarial \cite{feinman2017detecting} \cite{papernot2017extending}.
\par
Our model uses a combination defense: We base our model on Bayesian uncertainty in a dropout neural network \cite{srivastava2014dropout}, but use a secondary defense, preprocessing, as a double-check for \enquote{edge} cases. 
The existence of the secondary defense allows us to tune the uncertainty aspect of our defense in favor of declaring \enquote{edge} images (images close to the threshold uncertainty) adversarial, with less sacrifice of clean-image accuracy than would otherwise be the case. 


\section{Related Work}
Feinman et al. \cite{feinman2017detecting} proposed a defense based on Bayesian uncertainty with dropout, combined with kernel density estimation. 
They tested it with good results against adversarial examples generated via the Fast Gradient Sign Method (FGSM) \cite{goodfellow6572explaining}, Basic Iterative Method (BIM) \cite{kurakin2016adversarial}, Jacobian-based Saliency Map Attack (JSMA) \cite{papernot2016limitations}, and C\&W \cite{carlini2017adversarial} attacks. 
They achieved ROC-AUC scores on sets of adversarial images generated from the MNIST dataset ranging from 90.57\% to 98.13\% depending on attack method. 
This approach took advantage of the uncertainty estimates possible with dropout networks by assuming that the Bayesian uncertainty will be greater for adversarial examples than for clean data, because of the effect of the randomization on the necessarily precise perturbations. 
In addition, they used a Gaussian Mixture Model to analyze the outputs of the last hidden layer of their neural network, arguing that adversarial images will have a different distribution than clean ones. 
They also incorporated a kernel density estimate defense and evaluated their approach on MNIST \cite{lecun1998mnist}, CIFAR-10 \cite{krizhevsky2009learning}, and SVHN \cite{netzer2011reading} datasets.
\par
Papernot and McDaniel \cite{papernot2017extending} also presented an uncertainty based defense: A model-agnostic system in which the uncertainty is based on the predictions of a second, separate neural network which is used to train the classification network. 
This defense showed good results for the MNIST dataset, tested with adversarial images generated by FGSM, JSMA, and AdaDelta \cite{carlini2017towards} attacks.
\par
Several researchers have considered preprocessing based defenses against adversarial examples. 
Graese, Rozsa, and Boult \cite{graese2016assessing} explored several preprocessing techniques with the MNIST dataset, tested with FGSM and Fast Gradient Value (FGV) \cite{rozsa2016adversarial} attacks, and found the best results with cropping and resizing: 76\% and 78\% accuracy for FGV and FGSM samples respectively, compared to 65\% and 68\% for the next best \enquote{translation} technique on their raw model. 
Guo et al. \cite{guo2017countering} tested cropping-resizing as well as other image transformations (image quilting, JPEG compression, etc.) and found that cropping-resizing was \enquote{very efficient} with accuracy up to 76\% depending on the strength and method of attack.


\section{Methodology}
Our method uses Bayesian uncertainty \cite{gal2016dropout} \cite{feinman2017detecting} with a relatively low \enquote{adversarial} threshold for initial assessments and corrects for false negatives using image pre-processing.
Our method involves a Convolutional Neural Network (CNN) with dropout which reports the Bayesian uncertainty of its classifications, but which then takes multiple crops of images which fall near our threshold uncertainty level and classifies each crop again using our CNN. 
The final classification or adversarial image indication is based on both the level of Bayesian uncertainty and the agreement or lack thereof among the crops. 
Algorithm \ref{Algo} illustrates the procedure.
\begin{algorithm}[t]
\SetKwInOut{Input}{input}
\SetKwInOut{Output}{output}
\Input{an image $img$, a high threshold $H$, a low threshold $L$, and a count of agreed predictions $C$.}
\Output{image label (clean or adversarial).}
calculate $img$ prediction\;
calculate $img$ uncertainty\;
\eIf{$uncertainty > H$}
{declare adversarial\;}
{
\eIf{$uncertainty < L$}
{declare clean\;}
{
\For{1:5}{
crop \& resize $img$\;
classify $img$\;
\If{class = prediction}
{increment counter\;}
}
\eIf{$counter > C $}
{declare clean\;}
{declare adversarial\;}
}
}
\caption{Detecting adversarial images during inferring time.}
\label{Algo}
\end{algorithm}

\subsection{Computing Model Bayesian Uncertainty}
Dropout was first introduced as a means of avoiding overfitting in deep neural network training \cite{srivastava2014dropout}. 
Dropout layers mean that some weights are randomly zeroed out, that is, some links between neurons are cut. 
Gal and Ghahramani \cite{gal2016dropout} noted that including dropout layers in a neural network provides information about uncertainty for a wide variety of DNN architectures, sometimes without modification. 
Srivastava et al. \cite{srivastava2014dropout} used dropout at training time, however Gal and Ghahramani applied dropout to the inference stage.
\par
Our model uses dropout layers in the inference stage: For each image, we made $N$ stochastic passes through the network. 
Each pass produced a probability for each available class by applying Softmax to the resulting logit vectors $z_{1}(x), ..., z_{N}(x)$. 
\par
To obtain the stochastic prediction, we take the mean of the logit vector $z(x)$ for each class. 
The image is provisionally classified as the class with the highest mean. 
We also obtain the stochastic uncertainty by computing the standard deviation of the predictions over the $N$ stochastic passes.

\subsection{Image Preprocessing}
At this point, if the stochastic uncertainty is sufficiently high, our model declares the image adversarial without reference to the secondary method. 
If the uncertainty is sufficiently low, our model outputs the provisional classification as the final classification, again without reference to the secondary method. 
However, if the uncertainty lies within a certain middle range, the model employs the crop-resize secondary method.
\par
In the cropping stage, the network takes any image which has been classified with an uncertainty laying in the \enquote{edge} range, splits it into five overlapping crops (top left, bottom left, top right, bottom right, and center), in the manner of Graese et al. \cite{graese2016assessing}, and resizes each crop back to the original size. 
Each of these crops is reclassified by the network, and if four out of the five new classifications agree with the original, the image is not declared adversarial and the original classification is output. 
This allows us to set the \enquote{threshold} Bayesian uncertainty level low enough to risk a noticeable number of false negatives (clean images declared adversarial) at the uncertainty stage of the process, while providing a mechanism to prevent these \enquote{extra} false negatives from making a large impact on our model's final accuracy. 
The lower uncertainty threshold level, in turn, gives a greater chance of detecting borderline adversarial images. 


\section{Experiments}
We implemented our model using Python 3.6 with CleverHans 2.0, TensorFlow, Keras 2.0, Scikit-learn, OpenCV, and Pillow 5.1.0. 
CleverHans facilitates the generation of adversarial examples using a variety of techniques. 
Using multiple techniques to generate a set of adversarial images for testing allows us to assess the general applicability of our defense. 
We trained and tested our model using the MNIST dataset: a dataset of handwritten digits consisting of 60,000 training examples and 10,000 test images. 
We tested our model with 50/50 mixed clean and adversarial test sets for each of three attacks.

\subsection{Neural Networks}
For MNIST, we used the LeNet \cite{lecun1989backpropagation} convnet architecture. 
We used a dropout rate of 0.5 after the last pooling layer and after the inner-product layer, as in \cite{feinman2017detecting}. 
On normal, non-adversarial samples, this network shows an accuracy of 98\%.

\subsection{Adversarial Attacks}
We evaluated our model using three different methods of generating adversarial images: FSGM \cite{goodfellow6572explaining}, BIM \cite{kurakin2016adversarial}, and JSMA \cite{papernot2016limitations}.



\begin{center}
\begin{table}[h!]
\center
\caption{Classification accuracy of undefended model.}
\begin{tabular}{ m{3.5cm} | m{2cm}| m{2cm} | m{2cm}} 
\hline
 & FGSM & JSMA & BIM \\
\hline
Classification accuracy & 59.0\% & 52.0\% & 50.0\% \\
\hline
\end{tabular}
\label{undefended}
\end{table}
\end{center}

\begin{center}
\begin{table}[h!]
\center
\caption{Results for adversarial image detection.}
\begin{tabular}{ m{3.5cm} | m{2cm}| m{2cm} | m{2cm}} 
\hline
 &  FGSM & JSMA & BIM \\
\hline
False negative & 252 & 85 & 249 \\
\hline
False positive & 134 & 74 &104 \\
\hline
True negative & 4796 & 4856 & 4826 \\
\hline
True positive & 4725 & 4892 & 4728\\
\hline
Detection accuracy & 96.1\% & 98.3\% & 96.4\% \\
\hline
\end{tabular}
\label{detection}
\end{table}
\end{center}

\begin{center}
\begin{table}[h!]
\center
\caption{Results of applying the defense with different attacks using MNIST dataset.}
\begin{tabular}{ m{1.5cm} | m{1.7cm} | m{1.7cm} | m{1.7cm} | m{1.7cm} | m{2cm} } 
\hline
\multirow{2}{1.1cm}{Attack}
& 
\multicolumn{2}{|c|}{Clean images reported clean}
&
\multicolumn{2}{|c|}{Adv. images reported clean}
& 
\multirow{2}{1.1cm}{Classification Accuracy}
\\ \cline{2-5}
& Classified correctly & Classified incorrectly & Classified correctly & Classified incorrectly & \\ 
\hline
FGSM  &  97.3\% & 0.0\% & 1.7\% & 1.0\% & 99.0\% \\
\hline
JSMA  & 98.5\% & 0.0\% & 1.4\% & 0.1\% & 99.9\% \\
\hline
BIM  & 97.8\% & 0.0\% & 1.2\% & 1.0\% & 99.0\% \\
\hline
\end{tabular}
\label{result}
\end{table}
\end{center}

\section{Results}
For our model, we adjusted the uncertainty levels based on the information gain to improve accuracy and used these thresholds to calculate high and low levels of Bayesian uncertainty. 
Our experiments also showed that requiring four out of five crops be in agreement produced the greatest ultimate accuracy. 
We generated adversarial images using FGSM, BIM, and JSMA, with separate test runs for each attack.
\par
We tested our model with a set of 10,000 images comprised of 50\% clean and 50\% adversarial images. 
After testing with the basic, undefended CNN, we removed clean images which were misclassified; in testing the defended model, we used only adversarial images generated from images which were correctly classified by the basic CNN when clean.
\par
For the FGSM attack, as seen in Table \ref{undefended}, our basic CNN without defense had an accuracy of 59\% on the 50/50 mixed test set; most of adversarial images were indeed misclassified.  
By adding our defense to the network, we were able to detect adversarial images with an accuracy of 96.1\% which increased classification accuracy, for the images flagged as clean, to 99.0\% (see Table \ref{detection} and \ref{result}).
\par
Adversarial images produced using the BIM attack were even more successful against the unprotected CNN, which was only 50\% accurate on the mixed test set. 
With our defense added, our model handled a BIM mixed test set with an accuracy detecting adversarial images of 96.4\% and a classification accuracy on clean-flagged images of 99.0\%.
\par
On a mixed test set produced with the JSMA attack, our basic CNN had an accuracy of 52\%. 
With our defense, our model achieved 99.9\% classification accuracy with 98.3\% detection accuracy.
\par
In all cases above, accuracy was calculated as \textless images correctly classified\textgreater/ \textless total remaining (flagged as clean) images\textgreater, and did not include images discarded as adversarial. 
Detection accuracy was calculated using the \textless true positive + true negative\textgreater/\textless total images in set\textgreater. Total images now are less than 10,000 due to removal of misclassified clean images from consideration.
\par
Feinman et al. \cite{feinman2017detecting} used a Bayesian uncertainty model, but combined it with kernel density rather than preprocessing. 
With MNIST, they achieved mixed test set adversarial sample detection ROC-AUC scores of 90.57\% for FGSM, 97.23\% for BIM-A, and 98.13\% for JSMA, slightly lower than the detection accuracy results for our model. 
On average, Feinman et al. achieved an MNIST detection ROC-AUC of 95.31\%, somewhat lower than our average MNIST detection accuracy of \textbf{97\%}.
\par 
Graese, Rozsa, and Boult \cite{graese2016assessing} tested their techniques with the FGSM attack, but did not consider BIM nor JSMA attacks. 
For their 5-crops-and-resize method they reported a classification accuracy of 90.94\% for FGSM, noticeably lower than our combined model results of \textbf{99.0\%} for that attack, however their binarization defense had a comparable accuracy of 99.21\%. 
\par 
Wang et al. \cite{wang2018detecting} applied a set of mutations, obtained by imposing a minor random but realistic perturbation on the image. 
Based on their observation, clean images preserved their labels while adversarial did not. 
They scored, on average, 88.0\% for MNIST detection. Ma et al. \cite{ma2018characterizing} used Local Intrinsic Dimensionality (LID) to characterize the dimensional properties of adversarial subspaces or adversarial regions which facilitates the distinction of adversarial examples. For MNIST, on average, they scored 96.20\% accuracy, slightly lower than our average.


\section{Conclusion}
Our combined Bayesian uncertainty and image pre-processing proved effective, with accuracy in the high 90s, at detecting adversarial examples in a mixed MNIST test set.
This allowed for considerably higher accuracy in classifying the remaining, primarily clean, images. 
Our model has shown results which are for the most part similar to those of other recent work, with slightly greater accuracy in adversarial example detection for the test sets and attacks we considered. 
\par
Future work in those areas could be promising, as our results suggest that a combined defense has some advantages. 
In particular, it would be interesting to explore a combination method that involves binarization in the manner of Graese et al.  
Also, future work in defenses against additional attack methods could shed light on the general applicability of this combined method.


\bibliographystyle{splncs03}
\bibliography{ref}

\end{document}